\documentclass{article}
\usepackage{ijcai21}

\usepackage{algorithm}
\usepackage{algorithmicx}
\usepackage{algpseudocode}
\usepackage{multirow}
\usepackage{times}
\usepackage{soul}
\usepackage{url}
\usepackage[hidelinks]{hyperref}

\usepackage[small]{caption}
\usepackage{graphicx}
\usepackage{amsmath}
\usepackage{amsthm}
\usepackage{algorithm}
\graphicspath{{figures/}}
\usepackage{latexsym}
\usepackage{soul}
\usepackage{amsmath}
\usepackage{thmtools}
\usepackage{thm-restate}

\usepackage[utf8]{inputenc} 
\usepackage[T1]{fontenc}    
\usepackage{url}            
\usepackage{booktabs}       
\usepackage{amsfonts}       
\usepackage{nicefrac}       
\usepackage{microtype}      
\usepackage{color}

\title{Momentum Gradient-based Untargeted Attack on Hypergraph Neural Networks}

\usepackage{amsthm}

\author{
Yang Chen$^1$\and
Stjepan Picek$^{2}$
\and
Zhonglin Ye$^1$\and
Zhaoyang Wang$^1$\and
Haixing Zhao$^1$
\\
\affiliations
$^1$School of Computer Science, Qinghai Normal University 810000, Xining, Qinghai, China\\
$^2$Radboud University, Houtlaan 4, 6525 XZ Nijmegen, Netherlands\\
\emails
chenyang2753@stu.qhnu.edu.cn,stjepan.picek@ru.nl, yezhonglin@qhnu.edu.cn,z.y.wang@stu.qhnu.edu.cn,hxzhao@qhnu.edu.cn
}

\begin{document}

\maketitle
\begin{abstract}
Hypergraph Neural Networks (HGNNs) have been successfully applied in various hypergraph-related tasks due to their excellent higher-order representation capabilities. Recent works have shown that deep learning models are vulnerable to adversarial attacks. Most studies on graph adversarial attacks have focused on Graph Neural Networks (GNNs), and the study of adversarial attacks on HGNNs remains largely unexplored. In this paper, we try to reduce this gap. We design a new HGNNs attack model for the untargeted attack, namely MGHGA, which focuses on modifying node features. We consider the process of HGNNs training and use a surrogate model to implement the attack before hypergraph modeling. Specifically, MGHGA consists of two parts: feature selection and feature modification.
We use a momentum gradient mechanism to choose the attack node features in the feature selection module. In the feature modification module, we use two feature generation approaches (direct modification and sign gradient) to enable MGHGA to be employed on discrete and continuous datasets. We conduct extensive experiments on five benchmark datasets to validate the attack performance of MGHGA in the node and the visual object classification tasks. The results show that MGHGA improves performance by an average of 2$\%$ compared to the than the baselines.
\end{abstract}

\section{Introduction}
Graph Neural Networks (GNNs) are widely used in tasks such as graph classification \cite{A1}, node classification \cite{A2, A3}, and link prediction \cite{A4,A5} due to their efficient learning ability and generalization capability. Graphs provide a useful way to represent pairwise connections between objects in real-world networks. However, they may not fully capture the complex higher-order relationships between objects \cite{WOS:000934156000015}. When dealing with multimodal data, GNNs cannot efficiently learn all the information between objects, resulting in missing information and lower efficiency \cite{feng2019hypergraph}. For example, in the scientist collaboration network, researchers are abstracted as nodes and edges represent the paper collaboration relationship. Here, the common graph cannot represent the situation where multiple researchers work together on a paper \cite{inproceedings1}. The hypergraph can clearly represent this complex relationship. Specifically, the hyperedge (an edge in a hypergraph is called a hyperedge) represents the collaboration of a paper, and a hyperedge connecting $K$ nodes means that  $K$ researchers collaboratively work on a paper. Thus, the hypergraph has an advantage over the common graph in modeling complex relationships. Hypergraph Neural Networks (HGNNs) based on the hypergraph also outperform  GNNs in many tasks \cite{WOS:000982470300001,WOS:000922295100001,WOS:000965739400001}, especially in the field of network security \cite{WOS:000404505600010}.

In recent years, many works have demonstrated the vulnerability of GNNs to adversarial attacks, resulting in degraded performance \cite{Transferable,Nettack}. HGNNs, as an extension of hypergraph deep learning on graph data, also show vulnerability to graph adversarial attacks \cite{hu2023hyperattack}. Graph adversarial attacks aim to disrupt the performance of GNNs by adding small perturbations to the graph \cite{chen2022understanding,2022arXiv220812815L}. Depending on the goal of the attack, the adversarial attacks can be classified as targeted and untargeted attacks \cite{wang2022revisiting,liu2022towards}. In the targeted attack, the attacker focuses on the classification of some test nodes. The attack is successful only if the target node is misclassified to the attacker-specified label  \cite{dai2022targeted}. In the untargeted attack, the attacker usually focuses on the classification of all test nodes, and the attack succeeds if the test nodes are misclassified \cite{DBLPLinBW22}. Since the targeted attack usually attacks users with higher privileges, they are easily detected by defense models and are difficult to implement in real attacks \cite{Jin2020GraphSL}. Therefore, many works are based on the untargeted attack \cite{2022arXiE220801819T,2018arXiv181004714D}.

Almost all of the current works on graph adversarial learning focus on GNNs and ignore the security of HGNNs \cite{NIPA,AFGSM,9775013}, which makes HGNNs difficult to apply in practice. For example, adding some malicious noise to the pathology data can make it hard for doctors to understand the patient's condition and make wrong decisions in the task of detecting mental illness (e.g. Alzheimer's disease) \cite{WOS:000995531800001}. 

There are some differences between HGNNs and GNNs when dealing the data. For example, GNNs can only process traditional graph data, while HGNNs can not only process traditional graph data but also complex and high-dimensional data. Due to the specificity of hypergraph data, we summarize two main challenges in HGNNs attacks from the perspective of hypergraph data:

(1) \textbf{Unstructuredness}. Unlike common graph datasets, there is no association between nodes in most hypergraph datasets, and the hypergraph structure (adjacency relations) can be obtained through various modeling methods \cite{9264674}.

(2) \textbf{Continuous Features}. Many common graph datasets are consist of discrete features. Hypergraph datasets can be applied in tasks such as graph visualization \cite{WOS:000706330100042}, image classification \cite{WOS:000756892900005}, etc. Many hypergraph datasets are continuous features, so many attacks on GNNs cannot be applied to HGNNs \cite{9795251}.

\begin{figure*}[!h]
	\centering
	\includegraphics[width=0.7\textwidth]{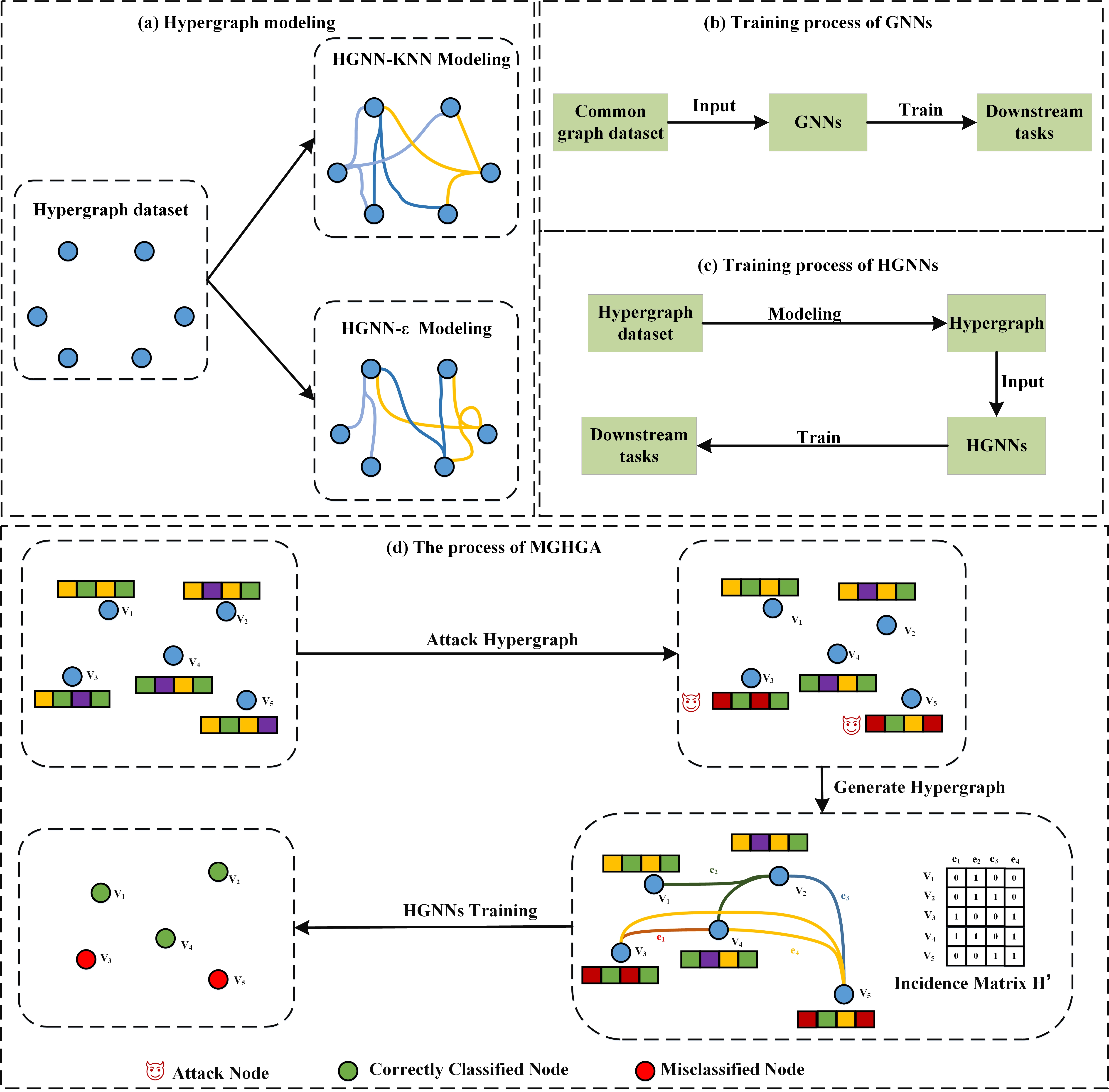}
	\caption{(a) shows two hypergraph modeling approaches. (b) and (c) represent the GNNs and HGNNs training processes, respectively. (d) shows the process of MGHGA.}
	\label{fig1}
\end{figure*}

HGNNs differ from GNNs in the convolution operation. Specifically, graph convolution is defined based on edges between nodes, while hypergraph convolution is defined based on hyperedges between nodes \cite{GCN,WOS000866212300048}. Hypergraph convolution is more complex than graph convolution, which leads to more difficult implementation of attacks in HGNNs. To the best of our knowledge, only HyperAttack \cite{hu2023hyperattack} has made a preliminary exploration of adversarial attacks for HGNNs. HyperAttack uses the gradient to modify the hyperedges. 
However, the HyperAttack implementation assumes that the hypergraph dataset has been modeled to obtain the hypergraph structure (the original hypergraph dataset is unstructured). 
In other words, HyperAttack implements the attack in the hypergraph with fixed structure.
Fig. \ref{fig1} (a) shows the hypergraph structure $H$ obtained using two distance-based HGNNs (HGNN-KNN, HGNN-$\varepsilon$ \cite{huang2009video}). We observe that the hypergraph structure $H$ is obtained differently under different approaches. In addition, different settings of HGNNs' parameters generate different hypergraph structures in practice. 
Therefore, HyperAttack has attack destabilizability as it exhibits different performances in different structural hypergraphs generated under the same dataset.
Intuitively, attacking hyperedges is not the best choice in HGNNs. This is because hypergraph datasets are unstructured and the same datasets generate the different hypergraph structures. The defender can learn from previous experience in adding or deleting many useless hyperedges to reduce the efficiency of the attack \cite{dai2018adversarial}.

To address the above challenges, in this paper, we propose an attack that is more applicable to HGNNs, namely MGHGA. MGHGA sets up the utility wider untargeted attack. Fig. \ref{fig1} (b) and (c) represent the training process of GNNs and HGNNs, respectively. We observe that modeling the dataset is a prerequisite for training HGNNs, whereas GNNs do not require this step. The previous attack algorithms directly modify the structure of the graph \cite{zhu2019robust,hussain2022adversarial}, and HyperAttack is no exception. To address the challenge (1), we take a new perspective of attacking features before hypergraph modeling. MGHGA uses momentum gradients to guide the attacker to modify the features. To address challenge (2), we consider the scenarios of attacking discrete and continuous features separately. In discrete datasets, MGHGA directly inverts the feature values from 0 to 1 or 1 to 0. In continuous datasets, we update the features using a sign gradient strategy. Finally, the new hypergraph structure is obtained by modeling the perturbed hypergraph dataset and feeding it into the HGNNs to verify the validity of the MGHGA. The above process is shown in Fig. \ref{fig1} (d). To highlight the innovations of our model, Table \ref{Comparison} shows the differences between HyperAttack and MGHGA.

\begin{table*}[h!]
	\centering
	\begin{center}
		\caption{Comparison of HyperAttack and MGHGA.}\label{Comparison}%
	
			\begin{tabular}{c|c|c|c}
				\toprule
				Model & Target of attack&Phase of attack & Downstream task\\
				\midrule
				HyperAttack& Targeted attack&After hypergraph modeling&Node classification\\
				MGHGA& Untargeted attack&Before hypergraph modeling&Node and Visual object classification\\
				\midrule
		\end{tabular}
	\end{center}
\end{table*}

Finally, we summarize the contributions of this paper as follows:

$\bullet$ We propose the first untargeted adversarial attack MGHGA against HGNNs. MGHGA considers the training characteristics of HGNNs and implements the attack before hypergraph modeling.

$\bullet$ We propose a momentum gradient method to guide the attacker in modifying the node's features, and our model can be applied to both  discrete and continuous hypergraph datasets.

$\bullet$ Extensive experiments on five datasets verified that MGHGA can effectively reduce the effectiveness of HGNNs and outperform other baseline methods in node and visual object classification tasks.

The rest of this work is organized as follows. In Section \ref{S3}, we introduce some fundamentals of the HGNNs and untargeted attack. Section \ref{S4} introduces MGHGA in detail, including feature selection and feature modification. Section \ref{S5} presents the experimental dataset, parameters and experimental results. In Section \ref{S2}, we first review the work related to hypergraph learning and graph adversarial attacks. Finally, we conclude our work in Section \ref{Sec:conclusion}.

\section{Preliminaries}\label{S3}
For convenience, Table \ref{notations} gives the frequently used notations.

\begin{table}[!ht]
	\begin{center}
		\caption{Notations frequently used in this paper and their corresponding descriptions.}\label{notations}%
		\begin{tabular}{cc}
			\toprule
			Notation & Description\\
			\midrule
			$\mathcal{D}$& Clean hypergraph dataset\\
			$\mathcal{D^{\prime}}$& Poisoned hypergraph dataset\\
			$\mathcal{G}$& Hypergraph\\
			$V$& Set of nodes of the clean hypergraph\\
			$X$& Feature matrix of the hypergraph graph\\
			$X^{\prime}$&  Feature matrix of the perturbed hypergraph\\
			$E$& Set of hyperedges of the clean hypergraph\\
			$W$& Hyperedge weight matrix\\
			$H$& Correlation matrix of the clean graph\\
			${D_e}$& Hyperedge degree \\
			${D_v}$& Node degree \\
			$Y$&True label\\
			$C$& Prediction label\\
			$F$& Feature gradient matrix\\
			$\Delta$ & Attack budget\\	
			$u$ &Momentum decay\\
			$ \eta $&Constraint factor \\
			$L_{model}( \cdot )$& HGNNs loss\\
			\midrule
		\end{tabular}
	\end{center}
\end{table}

\subsection{Hypergraph Neural Network}\label{3.1}
Given a hypergraph dataset $\mathcal{D} = (V,X)$,  where $V = \{ {v_1},{v_2},...,{v_{|V|}}\} $ represents the set of nodes, $X \in {\mathbb{R}^{|V| \times d}}$ denotes the node feature matrix and $d$ denotes the dimensions of the feature. Constructing the hypergraph $\mathcal{G} = (V,E,W)$, where $E = \{ {e_1},{e_2},...,{e_{|E|}}\} $ represents the set of hyperedges, and $diag(W) = [w({e_1}),w({e_2}),...,w({e_{|E|}})]$ is the diagonal matrix of the hyperedge weights, $w(e)$ is the weight of the hyperedge. The correlation matrix $H \in {\{ 0,1\} ^{|V| \times |E|}}$ is used to represent the structure of the hypergraph, $H(v,e) = 1$ if the node $v$ is inside the hyperedge $e$, and $H(v,e) = 0$ otherwise, which can be expressed as follows:
\begin{equation}
H(v,e) = \left\{ {\begin{array}{*{20}{c}}
	{1,\quad if \quad v \in e.}\\
	{0,\quad if \quad v \notin e.}
	\end{array}} \right.
\end{equation}

Due to the significant performance achieved by HGNNs widely used in classification \cite{huang2021residual}. In this paper, the downstream task of HGNNs is set as node classification. The hypergraph convolutional network learns node representations, transforms and propagates information to accomplish downstream tasks. A hypergraph convolutional layer can be represented as \cite{feng2019hypergraph}
\begin{equation}
X^{(l+1)}=\sigma\left(D_v^{-1 / 2} H W D_e^{-1} H^T D_v^{-1 / 2} X^{(l)} \Theta^{(l)}\right).
\end{equation}

where ${X^{(l)}}$ denotes the node representation of the hypergraph at $l$-th layer. $\sigma ( \cdot )$ denotes the nonlinear activation function, $W$ is the weight matrix of the hyperedges. ${D_e}$ represents the diagonal matrix of the hyperedge degree (hyperedge degree is the number of nodes contained in the hyperedge), ${D_v}$ is the diagonal matrix representing the node degree (node degree is the number of hyperedges containing the node), ${\Theta ^{(l)}}$ denotes the $l$ layer training parameters.

We set the layer number of HGNNs to 2 as in most other works \cite{feng2019hypergraph}, whose definition can be expressed as
\begin{equation}
Z=f(H, X)=\operatorname{softmax} \left(\widehat{H} \operatorname{Re} L U\left(\widehat{H} X \Theta^{(1)}\right) \Theta^{(2)}\right).
\end{equation}

where $\widehat{H}=D_v^{-1 / 2} H W D_e^{-1} H^T D_v^{-1 / 2}$. ${\Theta ^{(1)}}$  and ${\Theta ^{(2)}}$ are denoted as the training parameters of the first and second layers, respectively. In the training phase our goal is to continuously optimize $\Theta  = ({\Theta ^{(1)}},{\Theta ^{(2)}})$ to obtain the optimal classifier ${f_{{\Theta ^*}}}(H,X)$:

\begin{equation}\mathop {\min }\limits_\Theta  {L_{model}} =  - \sum\limits_{u \in {V_L}} {{Y_u} \ln {(Z_{u,:})}}. \end{equation}

where $Z_{u,:}$ denotes the set of predicted labeling probabilities for node $u$ , ${Y_u}$ denotes the true label of node $u$, ${V_L}$ denotes the training set of node, and the predicted label of node $u$ is denoted as
\begin{equation}
{L_{pre}} = \arg \max ({Z_{u,:}}).
\end{equation}

\subsection{Untargeted Attack}\label{3.2}

The aim of untargeted is to reduce the global classification performance of HGNNs. Given a budget $\Delta$ which represents the number of times the attacker modifies the feature entities. The untargeted attack can be expressed as
\begin{equation}\label{eq5}
\begin{aligned}
& \arg \max _{X^{\prime}} \sum_{v \in V_T} \mathbb{I} \left(Y_v \neq C_v\right), \\
& \text { s.t. } C=\arg \max f_{\Theta^*}\left(H, X^{\prime}\right), \Theta^*=\arg \min L_{\text {model } }\left(H, X^{\prime}\right), \left\|X-X^{\prime}\right\| \leq \Delta.
\end{aligned}
\end{equation}

Where $C$ is the set of node prediction labels, $V_T$ denotes the test set of nodes. $\mathbb{I}(x)$ is the indicator function. If x is true, $\mathbb{I}(x)$ returns 1, otherwise $\mathbb{I}(x)$ returns 0.

The main rationale for Eq. \ref{eq5} is that in a model with high training error is likely to generalize poorly on the test set as well.

\subsection{Threat Model}\label{3.3}
Depending on whether the attack occurs before or after the training of HGNNs, it can be categorized into the evasion attack and poisoning attack \cite{sharma2023node,shafahi2018poison}. 
Evasion attack is performed on the trained HGNNs, and the attacker cannot modify the parameters or structure of the HGNNs \cite{zhang2022projective}.
Poisoning attack is performed before the HGNNs are trained, and the attacker can insert perturbations in the training data to interfere with the training process of the HGNNs \cite{10.1145/3512527.3531373}.
Evasion and poisoning attacks occur during the testing and training of HGNNs \cite{wang2020evasion,sharma2023node}, respectively. In an evasion attack, the attacker's goal is to modify the links or features of the test nodes causing the nodes to misclassify \cite{fan2021reinforcement}. In the real attack, the attacker cannot access all the test data. For example, in an e-commerce recommendation system, graph deep learning models are used to predict the recommendation of a target item based on the sales records of existing goods. The attacker cannot modify the information of the competitor's goods, and the evasion attack is not applicable to this situation.
However, the poisoning attack can solve this situation. For example, the attacker uses the surrogate model to train the goods dataset.
The feedback from the surrogate model guides the attacker to modify the goods dataset. Eventually, the dataset with malicious perturbations is obtained \cite{10.1145/3567420}. 
When the graph deep models are run on the perturbed datasets, which learns the representation with malicious information, which decreases the recommendation rate of the target goods.

We focus on achieving HGNNs attack under poisoning attack, and the core idea is to use the surrogate model to attack to get the perturbed dataset before the HGNNs are trained.
We study the robustness of HGNNs, so the surrogate model is set to HGNNs.

In addition, MGHGA is a white-box attack, which requires constant feedback information (e.g., gradient or node prediction labels) from surrogate HGNNs.

\section{Momentum Gradient Hypergraph Attack}\label{S4}
\begin{figure*}[!h]
	\centering
	\includegraphics[width=1\textwidth]{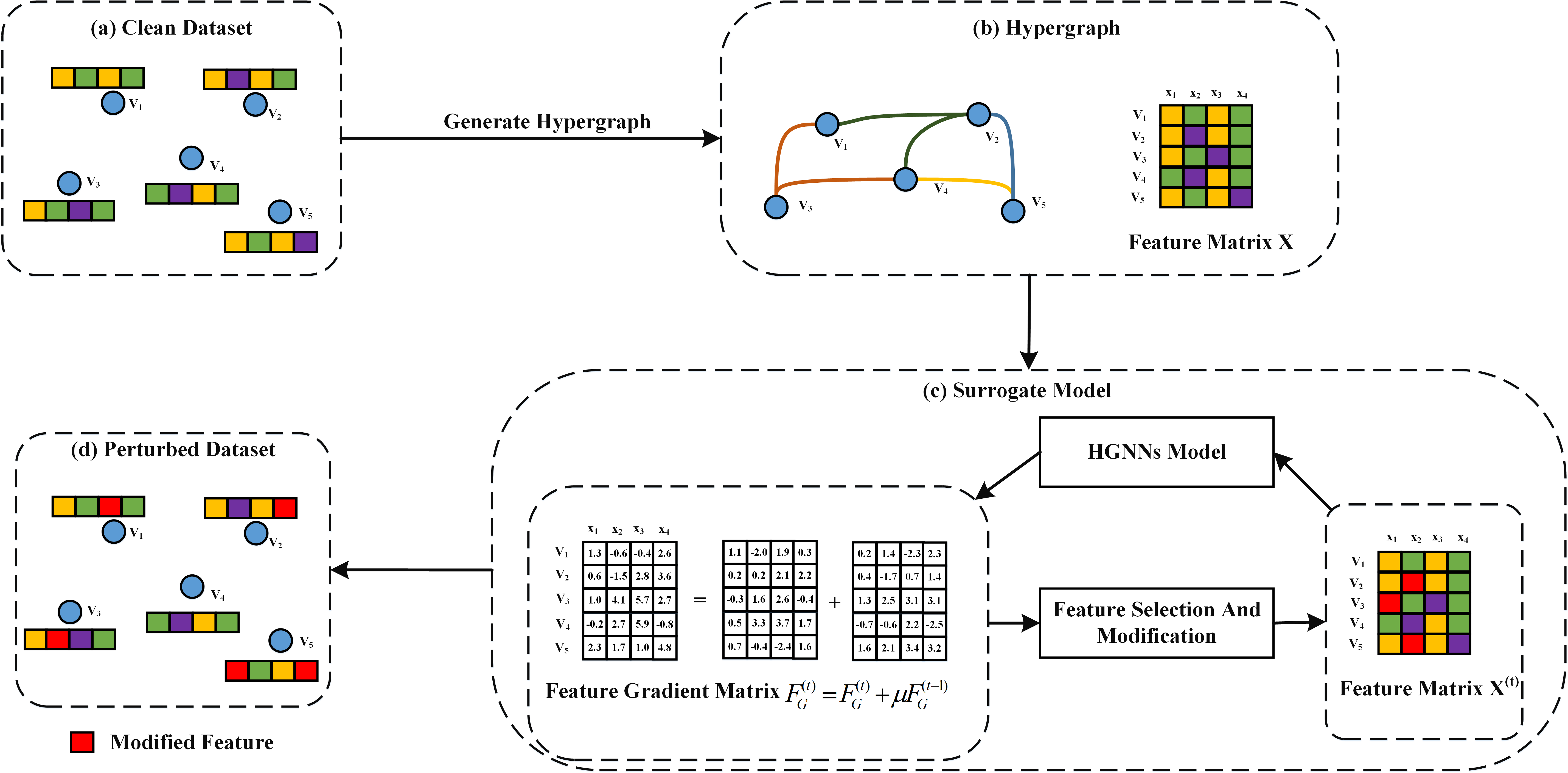}
	\caption{The pipeline of MGHGA. The goal of MGHGA is to generate perturbed datasets with malicious information. Where the red color represents the features of the nodes under attack.}\label{fig3}
\end{figure*}

In this section, we detail the MGHGA components. The MGHGA pipeline is shown in Fig. \ref{fig3}. MGHGA addresses two challenges in hypergraph attacks. First, many hypergraph datasets do not have correlation relationships between nodes. Attacking the hypergraph structure does not guarantee the stability of the attack, due to the different hypergraph structures generated for the same hypergraph dataset under different modeling approaches. To solve this problem, we use the surrogate model to attack the nodes' features before the hypergraph modeling, as shown in Fig. \ref{fig3} (c). Second, hypergraph datasets can be classified as discrete and continuous based on feature attributes. In order to improve the MGHGA, we design two methods to update the features.

\subsection{Feature Selection}\label{4.1}
We use the gradient of the feature matrix in the surrogate model to select features, denoted as

\begin{equation}\label{eq8}
{F_{i,j}} = \frac{{\partial {L_{model}}}}{{\partial {X_{i,j}}}}.
\end{equation}

Where, $F \in {\mathbb{R}^{|V| \times d}}$ is the gradient matrix of features. A larger gradient of a feature indicates that this feature has a greater impact on the optimization of HGNNs, and thus modifying the feature can often have a large impact on HGNNs. However, previous studies have shown that using a greedy approach to directly modify the feature with the largest gradient makes the attack susceptible to falling into the optimized local optimum and easily overfitting to the attack model, which can diminish the generalizability of the generated adversarial samples \cite{Nettack,2022arXiv220812815L}.

To address the above problem, we propose a momentum gradient hypergraph attack. The momentum method is a technique to accelerate the gradient descent algorithm by accumulating velocity vectors along the gradient direction of the loss function during the iteration \cite{chen2020mga}. Accumulating previous gradients helps the model avoid falling into local optimum. We apply the momentum method to generate malicious features. The momentum gradient matrix of the features is first computed.

\begin{equation}\label{eq9}
\left\{ {\begin{array}{*{20}{c}}
	{{F^0} = 0}.\\
	{{F^t} = \mu {F^{t - 1}} + \frac{{\partial {L_{model}}}}{{\partial {X^{t}}}}}.
	\end{array}} \right.
\end{equation}

where $u$ represents the momentum decay factor and ${X^{(t)}}$ denotes the perturbation feature matrix after $t$ iterations.

MGHGA select the feature of the maximum absolute value of the momentum gradient matrix.

\begin{equation}\label{eq10}
m_{i,j}^{t} = \arg \max |{F^{t}}|.
\end{equation}

where $m_{i,j}^{t}$ represents the absolute maximum value of gradient at $t$ iterations, $i$ and $j$ represent the node and feature indexs, respectively.

\subsection{Feature Modification}\label{4.2}
At each iteration, we modify only one feature and stop the attack when the number of modifications reaches the budget $\Delta$. Here, we consider discrete and continuous feature modifications, respectively.

\textbf{Discrete Feature}.
In discrete features, where the feature values are only 1 or 0, we use a direct inversion mechanism to modify the features. At $t$ iterations, feature deletion or addition can be expressed as

\begin{equation}\label{eq11}
X_{i,j}^t = 1 - X_{i,j}^t, \quad s.t. \quad  ||{X^t} - X|| \le \Delta .
\end{equation}

\textbf{Continuous Feature}. For the image attack with continuous features, the researchers use the gradient sign mechanism to generate perturbations, and the results indicate that the mechanism can achieve high efficiency attack in continuous data \cite{goodfellow2014explaining}. Inspired by this, we use a gradient sign mechanism to update the continuous features. Our approach involves introducing perturbations along the opposite direction of the normal sample gradient to maximize the target model's training loss error, thereby lowering its classification confidence and increasing the probability of inter-class confusion, ultimately leading to misclassification. Specifically, the update feature matrix can be expressed as

\begin{equation}\label{eq12}
X_{i,j}^t = X_{i,j}^t + \eta sign(F_{i,j}^t), \quad s.t. \quad ||{X^t} - X|| \le \Delta .
\end{equation}

where $ \eta $ is the constraint factor and $sign(x)$ is denoted as the gradient symbol. $sign(x) = 1$ when $x > 0$, otherwise $sign(x) = 0$.
At the end of the iteration, the anomalous features are filtered in order to avoid that the newly generated features are not within the original features.

\begin{equation}\label{eq13}
\left\{\begin{array}{l}
X_{i,j}^{\prime}=\arg \min X, \text { s.t. } X_{i j}^{\prime}<\arg \min X. \\
X_{i,j}^{\prime}=\arg \max X, \text { s.t. } X_{i j}^{\prime}>\arg \max X.
\end{array}\right.
\end{equation}

where $X$ and $X^{\prime}$ denote the original and perturbed hypergraph feature sets, respectively. $arg min(X)$ and $arg max(X)$ denote the minimum and maximum values of the original hypergraph feature set, respectively.

We reset the features that are beyond the lower and upper of the original features, which ensures the imperceptibility of the perturbations and prevents them from being detected by some simple defense mechanisms (e.g., outlier detection models). Note that filtering anomalous features in MGHGA does not consume budget.

Since multiple modifications of the same feature cause a waste of budget, we attack at most once for both continuous and discrete features.

\subsection{Algorithm}\label{4.3}
The pseudo-code of MGHGA is given in Algorithm \ref{alg1}.

\begin{algorithm}[ht!]  
	\renewcommand{\algorithmicrequire}{\textbf{Input:}}
	\renewcommand{\algorithmicensure}{\textbf{Output:}}
	\caption{MGHGA}  
	\label{alg1}
	\begin{algorithmic}[1] 
		\Require Hypergraph dataset $\mathcal{D} = (V,X)$, momentum decay factor $u$, constraint factor $\varepsilon $, budget $ \Delta $
		\Require Perturbation hypergraph dataset $\mathcal{D^{\prime}} = (V,X^{\prime})$
		
		\State \textbf{Initialization:} Modeling hypergraph  $\mathcal{G} = (V,E,W)$, number of attack iterations $T$, HGNN surrogate model ${f_{{\Theta ^*}}}(H,X)$, perturbation gradient matrix  ${F^0}$
		
		\While {$t<T$ or $||{X^t} - X|| \le \Delta $}
		\State Calculate the gradient of $i$ iterations through Eq. \ref{eq8}
		\State Calculate the $i$ iteration momentum gradient through Eq. \ref{eq9}
		\State Select the $i$ iteration features through Eq. \ref{eq10}
		
		\If{$X$ is the discrete feature}
		\State Update gradient matrix through Eq. \ref{eq11}
		\Else  \Comment{$X$ is the continuous feature}  
		\State Update gradient matrix through Eq. \ref{eq12}
		\EndIf
		\State $t=t+1$
		\EndWhile
		
		\If{$X$ is the continuous feature}
		\State Filter features through Eq. \ref{eq13}
		\EndIf
		
	\end{algorithmic}
\end{algorithm}

\textbf{Complexity Analysis.} MGHGA uses HGNN as a pre-training model, and HGNN includes forward and backward propagation in training with a complexity of $\mathcal{O}(t||H|| \cdot ||X||)$, where $t$ denotes the number of HGNN training. Then MGHGA calculates the gradient of the feature matrix, the complexity is $\mathcal{O}(d|V|)$. Updating and filtering features are basic operations with low complexity and are ignored here. The complexity of modifying features is $\mathcal{O}(Td|V|)$, where $T$ denotes the number of attack iterations. In summary the overall complexity of MGHGA is $\mathcal{O}(t||H|| \cdot ||X|| + Td|V|)$.

\section{Experiments}\label{S5}
\subsection{Datasets}\label{5.1}
Recent works have shown that HGNNs exhibit excellent performance on node classification and visual object classification tasks \cite{huang2021residual,feng2019hypergraph}, which are the most common practical applications of HGNNs. Therefore, our work focuses on these two tasks. To illustrate the performance of MGHGA, experiments are carried out on five datasets. We performed node classification tasks on Cora, Cora-ML and Citeseer \cite{sen2008collective} datasets. The visual object classification task is performed on two multi-feature and continuous datasets including National Taiwan University 3D model (NTU) \cite{chen2003visual} and Princeton ModelNet40 (ModelNet40) \cite{wu20153d}. The dataset information is summarized in Table \ref{tab1}. Hypergraphs are obtained by modeling hypergraph datasets. However, NTU and ModelNet40 are without adjacencies, and their adjacencies are obtained using the hypergraph construction methods. In order to ensure the consistency of the adjacency relations of each dataset, we do not use the original adjacency of the Cora, Cora-ML and Citeseer datasets but instead utilize commonly used construction methods to generate hypergraph structures in the experiments.

\begin{table*}[h]
	\begin{center}
		\caption{ Statistics of five datasets. We use two discrete datasets (Cora, Citeseer) and one continuous dataset (Cora-ML) in the node classification task, respectively. Visual object datasets are usually continuous datasets (NTU, ModelNet40), and we use two datasets to validate the model's effectiveness.}\label{tab1}%
		\begin{tabular}{cccccc}
			\toprule
			Datasets & $\#$ Nodes & $\#$ Features & $\#$ Classes &  $\#$ Binaries \\
			\midrule
			Cora     & 2485    & 1433    & 7       & Y       \\
			Cora-ML     & 2810    & 2879    & 7       & N       \\
			Citeseer & 3327    & 3703    & 6       & Y       \\
			\midrule
			Datasets & $\#$ Nodes & $\#$ Features1 & $\#$ Features2 & $\#$ Classes &  $\#$ Binaries \\
			\midrule
			NTU     & 2012    & 4096    & 2048       & 67 &N       \\
			ModelNet40     & 12311    & 4096    & 2048       & 40 & N       \\
			\midrule
		\end{tabular}
	\end{center}
\end{table*}

\subsection{Baselines}\label{5.2}
\textbf{Since MGHGA is the first work on the untargeted adversarial attack in HGNNs, there are fewer comparative models to refer}. HyperAttack, which is most relevant to our work, is set up as a targeted attack and cannot be used as a comparison model. Due to the specificity of the hypergraph structure, it is difficult to directly migrate the GNNs adversarial attack models to HGNNs. Here, we use the following model as comparison models.

\textbf{Random Attack (Random)}: The conclusion of the work on common graph attacks shows that the Random Attack can degrade the performance of GNNs \cite{Nettack}. In this paper, we attack the features randomly. Specifically, the features are randomly changed from 0 to 1 and from 1 to 0 in the discrete dataset. In the continuous dataset, the features are randomly modified. It should be noted that the modified features are within the range of the original features.

\textbf{Node Degree Attack (NDA)}: Previous works have shown that attacking nodes with maximum node degree degrades the performance of the GNNs \cite{wang2022fake}. Extending to HGNNs, 
NDA is a method to attack nodes with maximum node degree. Note that features are modified in the same way as Random.

\textbf{Fast Gradient Attack (FGA)}: Fast Gradient Attack is a common gradient attack in the common graph \cite{FGA}. We extend it to hypergraphs. In each attack, we choose the feature with the largest absolute value of the gradient to attack.

\textbf{Fast Gradient Attack-Node Degree (FGA-D)}: We add a constraint for the FGA that attacks the node with the larger degree, which obtains the FGA-D. Note that the FGA and FGA-D modify the discrete features similarly to MGHGA.

\textbf{MGHGA-D}: MGHGA-D is an extended model of MGHGA, and MGHGA-D is obtained from MGHGA with the same constraint as FGA-D.

Note that FGA-D and MGHGA-D are attacks with constraints, i.e., they attack nodes with larger node degrees.

\subsection{Parameter setting and metrics}\label{5.3}
\textbf{Parameters}. In our experiments, the hypergraph is generated using two distance-based generation methods, i.e., HGNN-KNN and HGNN-$\varepsilon$, where $K$ and $\varepsilon$ are set to 10 and 0.5, respectively. The correlation matrix $H$ is set to a binary matrix. HGNNs are set to two layers, the feature dimension of the hidden layer is set to 64 and dropout is applied to avoid overfitting. In the training process, the training count is set to 300 and the learning rate of the Adam optimizer is 0.001. The ratio of the training set to the test set is 0.2 and 0.8, respectively. In the discrete dataset, the constraint factor $\eta$  is set to ${X_{avg}}$. The constraint factor $\eta$ is set to ${X_{avg}} = \frac{{sum(X)}}{{|V|d}}$ in the continuous dataset. The budget $\Delta  = \lambda |V|$, where $\lambda $ is the budget factor is set to 0.05 by default. The decay factor $\mu$ is 0.8. FGA-D and MGHGA-D attack nodes with top 1$\%$ node degree by default.
The victim model and the target model are the same by default, where the victim model is the model used by the user and the victim model is the pre-trained model used by the surrogate model. The experiments are conducted on a computer with an Intel(R) Xeon(R) Gold 5118 processor and 2* NVIDIA GeForce GTX 1070Ti GPU.

\textbf{Metrics}. For a comprehensive evaluation of MGHGA, we use the classification success rate to measure the attack effectiveness. The classification success rate indicates the classification accuracy of HGNN in the test set, and a lower rate indicates a better attack.

\subsection{Experimental Results}\label{5.4}\
\subsubsection{MGHGA Attack Performance}\label{5.4.1}
Table \ref{tab2} summarizes the classification accuracies of the five types of datasets under the attacks. The performance of HGNN-KNN and HGNN-$\varepsilon$ decreases under all attacks, which indicates that attacking node features can effectively degrade the performance of HGNNs. Specifically,  we observe that MGHGA achieves the best performance under all the datasets. For example, using HGNN-KNN as the victim model, the classification accuracies of the discrete dataset Cora are 58.65$\%$, 58.47$\%$, 57.63$\%$ and 55.33$\%$ for Random, NDA, FGA and MGHGA, respectively. The lower classification accuracy indicates that the attack causes more damage to HGNNs. Therefore, MGHGA achieves the best efficiency in comparing the advanced attacks. The results are the same in other datasets, especially in NTU and ModelNet40 continuous datasets, which shows that our proposed method is applicable not only to discrete datasets but also to continuous datasets.

MGHGA improves the performance by 3$\%$ on average compared to Random.  In particular, MGHGA improves the performance by 5$\%$  in Citeseer, which shows that MGHGA can add some critically important perturbations with the same budget.
FGA shows significant performance on some new tasks due to its strong generalization ability. In the comparison models, the results of FGA can be viewed as the current optimum. With a small budget, our proposed model improves the attack performance by 2$\%$ on average, which is a satisfactory result for us.
HGNNs are better able to utilize global as well as longer range contextual information when aggregating neighboring features, resulting in improved robustness of HGNNs over GNNs. Attacking HGNNs is more difficult than attacking GNNs. MGHGA, as a preliminary exploration of targetless attacks on HGNNs, shows outstanding performance compared to all other models, which indicates that our proposed attack is capable of achieving an optimal attack.

In addition, we investigate the effect of adding attack constraints (attacking the node with the largest node degree) on the attacks. Comparing MGHGA-D and MGHGA found that the performance of MGHGA-D attacking the node with the largest node degree, although it can reduce the performance of GNNs, the attack performance is not as good as that of MGHGA without constraints.
For example, in Citeseer, the performance of MGHGA over MGHGA-D is improved by 2.08$\%$ and 2.67$\%$ in HGNN-KNN and HGNN-$\varepsilon$ respectively.
The same rule is exhibited in FGA-D and FGA.
Intuitively, the unrestricted attack can maximize the efficiency of the attack.

\begin{table*}[!ht]
	\caption{Comparison of classification accuracy ($\%$) of several attack models. The lower the classification success rate, the better the model performance. In each case, the best results are bolded. The results are the average of 10 runs.}\label{tab2}%
	\begin{center}
		\resizebox{\linewidth}{!}{
			\begin{tabular}{c|c|c|cccccc}
				\toprule
				Datasets & Model  & Clean & Random& NDA& FGA-D &FGA & MGHGA-D & MGHGA  \\
				\midrule
				\multirow{2}*{Cora} & HGNN-KNN &59.31$\pm$0.3&58.65$\pm$0.9&58.47$\pm$0.5&58.24$\pm$0.8&57.63$\pm$1.1& 58.13$\pm$0.9&\textbf{55.33$\pm$1.8}\\
				& HGNN-$\varepsilon$ &57.19$\pm$0.2&56.97$\pm$1.1&56.73$\pm$0.6&56.46$\pm$0.7&55.10$\pm$1.0& 54.94$\pm$0.9&\textbf{53.51$\pm$1.5}\\
				\midrule
				\multirow{2}*{Cora\_ML} & HGNN-KNN &69.33$\pm$0.2&68.89$\pm$0.8&68.34$\pm$0.6&68.11$\pm$0.5&67.76$\pm$1.0& 67.90$\pm$0.7&\textbf{66.12$\pm$1.2}\\
				& HGNN-$\varepsilon$ &69.13$\pm$0.3&68.77$\pm$0.8&68.19$\pm$0.4&68.04$\pm$0.8&67.42$\pm$0.8& 67.64$\pm$0.9&\textbf{65.76$\pm$1.3}\\
				\midrule
				\multirow{2}*{Citeseer} & HGNN-KNN &64.63$\pm$0.2&63.34$\pm$0.4&63.03$\pm$0.6&62.90$\pm$0.6&62.44$\pm$1.6& 60.13$\pm$0.7&\textbf{58.05$\pm$0.9}\\
				& HGNN-$\varepsilon$ &62.12$\pm$0.1&61.41$\pm$0.5&60.90$\pm$0.4&60.21$\pm$0.4&59.13$\pm$1.3& 59.34$\pm$0.7&\textbf{57.67$\pm$1.5}\\
				\midrule
				\multirow{2}*{NTU} & HGNN-KNN &75.06$\pm$0.2&73.99$\pm$0.4&73.23$\pm$0.5&72.92$\pm$0.6&72.11$\pm$1.0& 72.17$\pm$0.5&\textbf{71.22$\pm$1.2}\\
				& HGNN-$\varepsilon$ &73.73$\pm$0.2&73.06$\pm$0.7&72.61$\pm$0.6&72.13$\pm$0.5&70.30$\pm$0.8& 70.49$\pm$0.7&\textbf{69.18$\pm$1.4}\\
				\midrule
				\multirow{2}*{ModelNet40} & HGNN-KNN &89.91$\pm$0.2&88.15$\pm$0.4&87.93$\pm$0.5&87.54$\pm$0.7&86.67$\pm$0.8& 86.97$\pm$0.6&\textbf{85.64$\pm$0.9}\\
				& HGNN-$\varepsilon$ &88.12$\pm$0.3&87.49$\pm$0.4&87.02$\pm$0.4&86.46$\pm$0.5&85.82$\pm$0.9&86.07$\pm$0.6 &\textbf{84.19$\pm$1.3}\\
				\midrule
		\end{tabular}}
	\end{center}
\end{table*}

\subsubsection{Running Time}\label{5.4.2}
Table \ref{tab3} shows the runtimes for several attacks. Specifically, Random has a low runtime in each dataset, but its performance is the worst of all the attacks and therefore would not be considered for application in the real attack. Comparing FGA and MGHGA shows that MGHGA runtime can be similar to FGA, but enables more efficient attacks. For example, in Citeseer, the running times of FGA and MGHGA are 4.10 and 4.15 minutes, respectively. Their classification accuracies are 62.44$\%$ and 58.05$\%$ (obtained from Table \ref{tab2}) when the victim model is HGNN-KNN, respectively. Our results show that our model achieves efficient attack while also achieving runtimes similar to comparison models. In most cases, we find that the runtime of the attack is positively correlated with the size of the nodes, and the runtime is longer when the node size is larger. For example, the number of nodes in descending order are ModelNet40, Citeseer, Cora\_ML, Cora and NTU, and the runtimes in descending order are: ModelNet40, Citeseer, Cora\_ML, NTU and Cora. An exception exists where NTU$>$ Cora. Intuitively, NTU is two-featured data, and HGNNs spend longer time processing two-featured data than single-featured data.

\begin{table*}[!ht]
	\caption{Running time of various attacks in minutes.}\label{tab3}%
	\begin{center}
		\resizebox{\linewidth}{!}{
			\begin{tabular}{c|c|cccccc}
				\toprule
				Datasets & Model   & Random&NDA & FGA-D &FGA & MGHGA-D & MGHGA  \\
				\midrule
				\multirow{2}*{Cora} & HGNN-KNN &0.25&0.37&2.35&2.33&2.35&2.34\\
				& HGNN-$\varepsilon$ &0.25&0.37&2.35&2.33&2.35&2.34\\
				\midrule
				\multirow{2}*{Cora\_ML} & HGNN-KNN &0.25&0.35&3.33&3.31&3.34&3.33\\
				& HGNN-$\varepsilon$ &0.25&0.36&3.34&3.31&3.35&3.33\\
				\midrule
				\multirow{2}*{Citeseer} & HGNN-KNN &0.27&0.40&4.15&4.10&4.19& 4.15\\
				& HGNN-$\varepsilon$ &0.27&0.39&4.16&4.11&4.19&4.16\\
				\midrule
				\multirow{2}*{NTU} & HGNN-KNN &0.25&0.44&2.56&2.50&2.56&2.55\\
				& HGNN-$\varepsilon$ &0.25&0.44&2.56&2.49&2.56&2.55\\
				\midrule
				\multirow{2}*{ModelNet40} & HGNN-KNN &1.23&2.54&10.64&11.13&10.67&11.21\\
				& HGNN-$\varepsilon$ &1.26&2.58&10.66&11.16&10.71&11.18\\
				\midrule
		\end{tabular}}
	\end{center}
\end{table*}

\subsubsection{Attack performance in different budgets}\label{5.4.3}
\begin{figure*}[hpt]
	\centering
	\includegraphics[width=1\textwidth]{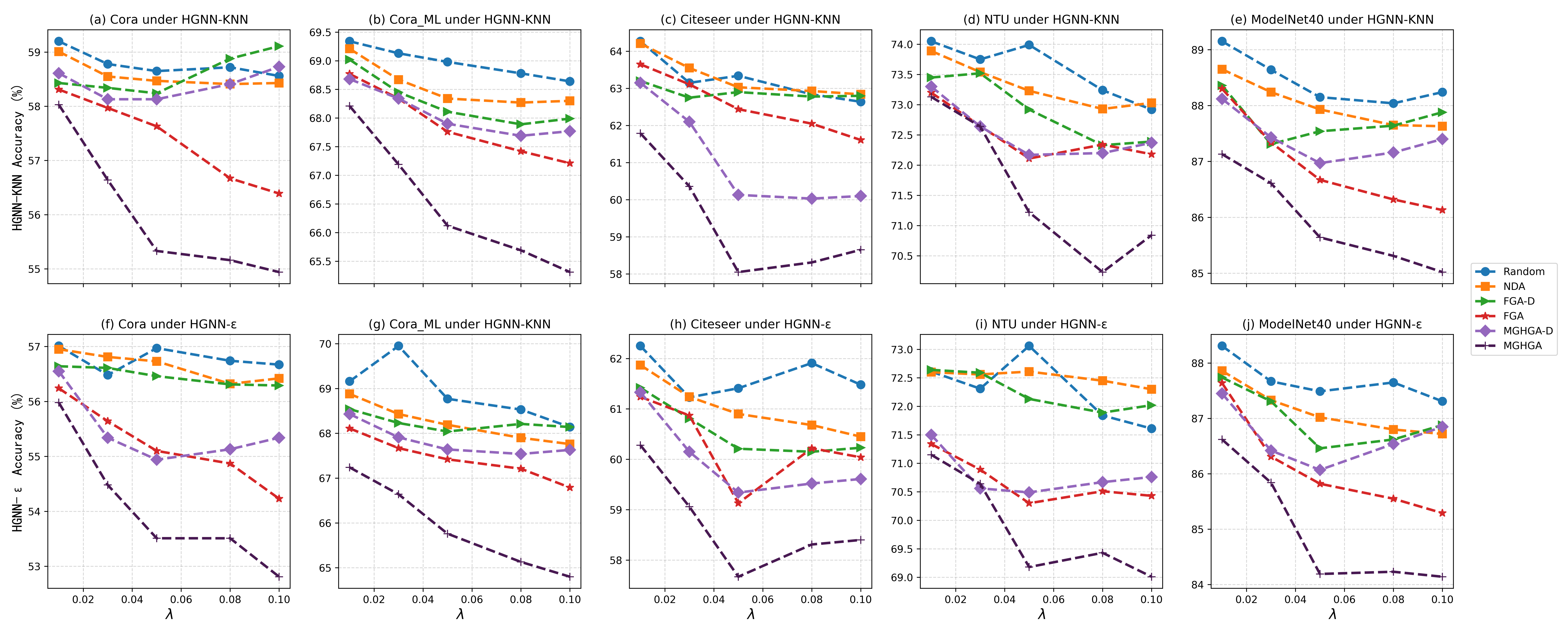}
	\caption{The attack performance under different budget factors $\lambda$.}\label{fig4}
\end{figure*}

We further investigate the performance of various attacks under different budgets to verify the effectiveness of MGHGA. As shown in Fig. \ref{fig4}, we observe that MGHGA achieves satisfactory results with different budgets in all datasets. In Fig. \ref{fig4} (j), the classification accuracies of FGA and MGHGA are $\{$87.64$\%$, 85.82$\%$, 85.29$\%$$\}$ and $\{$86.62$\%$, 84.19$\%$, 84.14$\%$$\}$ when the budget factor $\lambda$ are $\{$0.01, 0.05, 0.1$\}$, respectively. 

Furthermore, we investigated the effect of the budget on two constrained attacks (FGA-D and MGHGA-D).
Fig. \ref{fig4} shows that the increase in budget negatively affects the constrained attacks.
As an example, the classification accuracies of FGA-D and MGHGA-D are $\{$88.36$\%$, 87.54$\%$, 87.88$\%$$\}$ and $\{$88.12$\%$, 86.97$\%$, 87.40$\%$$\}$ in Fig. \ref{fig4} (e), when the budget factor $\lambda$ are $ \{ 0.01, 0.05, 0.1\} $. We analyze that the larger nodes (FGA-D and MGHGA-D attack the node with the larger node degree each time) have a limited impact on the attack performance. FGA-D and MGHGA-D attack features do not positively impact the attack when the budget is too large. Therefore, the performance of the restricted attack increases and then decreases with the budget increases, which is particularly evident in the Cora and ModelNet40 datasets.

\subsubsection{Attack performance in HGNNs parameters $K$ and $\varepsilon$}\label{5.4.4}
\begin{figure*}[hpt]
	\centering
	\includegraphics[width=1\textwidth]{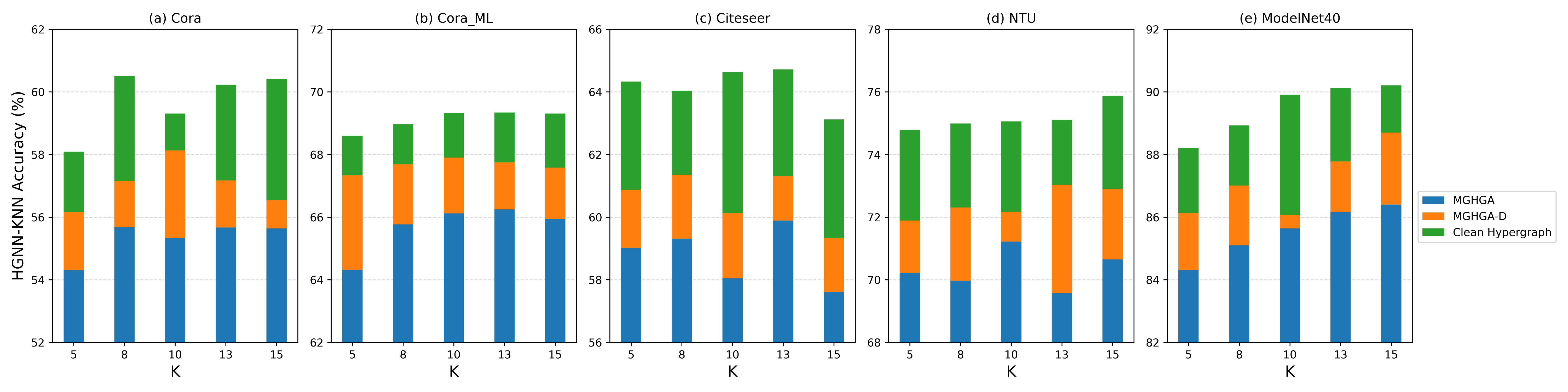}
	\caption{The model performance in different parameters $K$.}\label{fig5}
\end{figure*}

Fig. \ref{fig5} shows the performance of our proposed model for different parameter $K$. We find that the performance of MGHGA is independent of the victim model HGNN-KNN parameter $K$, i.e., MGHGA still reduces the accuracy of HGNN-KNN regardless of $K$. For example, in NTU, MGHGA-D and MGHGA reduce the accuracy of $\{$2.08$\%$, 3.84$\%$, 1.51$\%$$\}$, and $\{$3.90$\%$, 4.27$\%$, 3.81$\%$$\}$, respectively, when $K$ are $\{$5, 10, 15$\}$.

\begin{figure*}[hpt]
	\centering
	\includegraphics[width=1\textwidth]{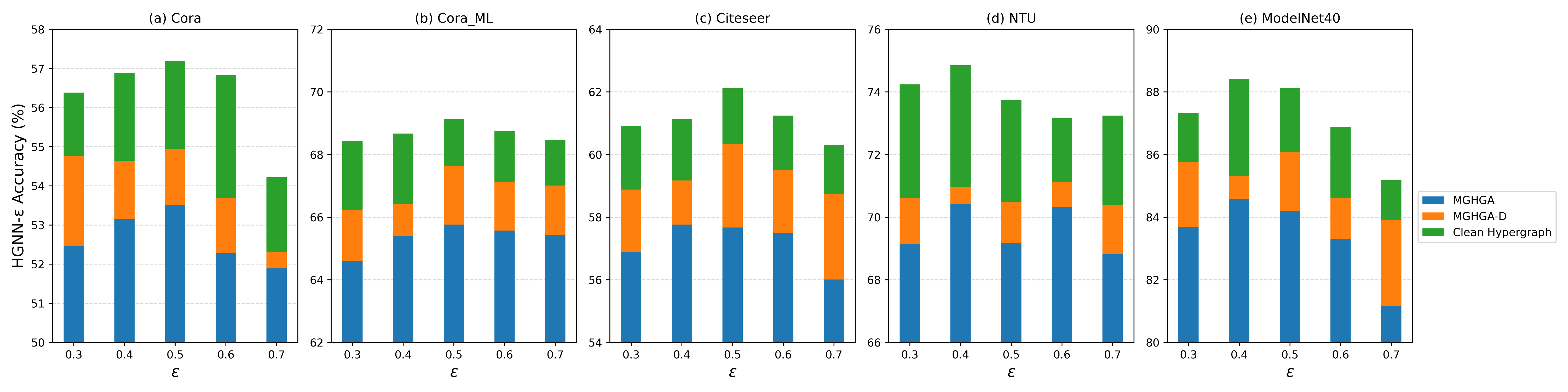}
	\caption{The model performance in different parameters $\varepsilon$.}\label{fig6}
\end{figure*}

We investigate the effect of parameter $\varepsilon$ on our model, and the results are shown in Fig. \ref{fig6}. In each dataset, MGHGA is able to achieve reduced HGNN-$\varepsilon$ classification accuracy. Specifically, the average performance of MGHGA-D and MGHGA in Citeseer is 3.03$\%$ and 4.43$\%$ under all $\varepsilon$, respectively.

The above results indicate that our model is not affected by the victim model parameters. Previous work has shown that the parameters $K$ and $\varepsilon$ do not affect the performance of HGNNs on classification tasks, i.e., HGNNs have good stability \cite{feng2019hypergraph}. We think that the parameters $K$ and $\varepsilon$ do not affect our model due to the stability of HGNNs.

\subsubsection{Attack performance in different decay factors}\label{5.4.5}

\begin{figure*}[hpt]
	\centering
	\includegraphics[width=1\textwidth]{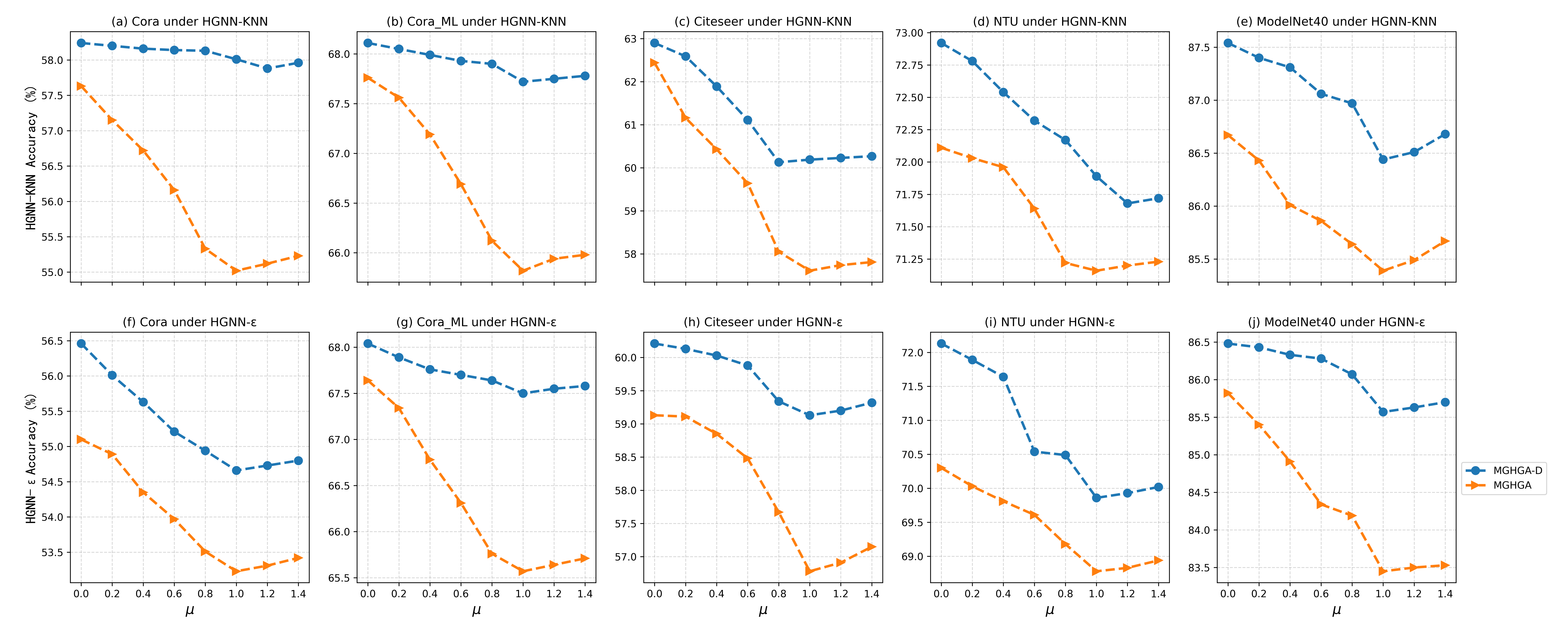}
	\caption{The model performance in different decay factor $\mu$.}\label{fig7}
\end{figure*}

The decay factor $\mu$ is an important parameter of our model. Fig. \ref{fig7} shows the effect of the decay factor $\mu$ on our model. Note that the MGHGA degrades to FGA when $\mu$= 0. We observe that the model performance first increases and reaches a plateau or decreases as $\mu$ increases. 
For example, in Fig.  \ref{fig7} (d), the accuracy of MGHGA are $\{$72.11$\%$, 71.21$\%$, 71.23$\%$
$\}$ when $\mu$ are $\{$0, 1, 1.4$\}$, respectively. 
In addition, our model performs best when $\mu$ = 1 for discrete and continuous data sets in most cases. Intuitively, when $\mu$ is small, the momentum gradient mainly depends on the gradient of the HGNNs at the $t$ moment, and the performance of MGHGA is similar to that of the normal gradient attack FGA. As $\mu$ increases, the gradient of the MGHGA depends on the gradients of the previous moments and the current moment , and this way of combining the gradients starts to improve the performance of the MGHGA. However, as $\mu$ continues to increase, the momentum gradient relies heavily on the gradient of the previous moments and ignores the feedback from the gradient of the latest moment, and MGHGA performance degrades. The above results indicate that using the momentum gradient model can improve the attack performance.

\subsubsection{Transferability of MGHGA}\label{5.4.6}

\begin{figure*}[hpt]
	\centering
	\includegraphics[width=1\textwidth]{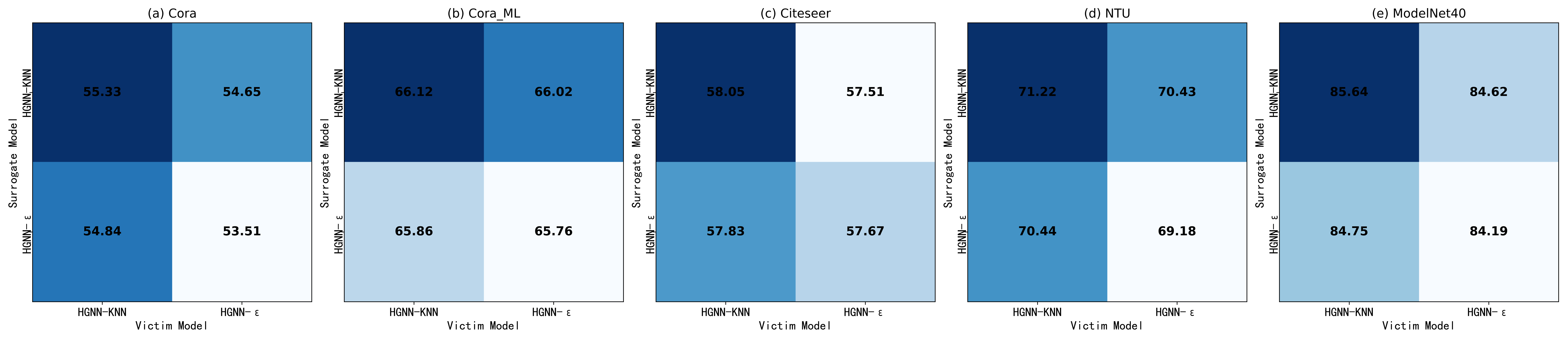}
	\caption{The translatability of MGHGA. Where the x-axis represents the victim model and the y-axis represents the surrogate model.}\label{fig8}
\end{figure*}

In this section, we verify the transferability of our model. The heat map \ref{fig8} illustrates the performance of MGHGA in various surrogate and victim models. It is observed that the combination of surrogate and victim models does not affect the MGHGA performance. Specifically, in Citeseer, the performance of IMGIA improved by 0.16$\%$ when the surrogate model and victim models are HGNN-KNN and HGNN-$\varepsilon$, respectively. We think that MGHGA completes the attack before the victim HGNNs are trained and do not need to access the training parameters of the victim model, so the choice of surrogate and victim models does not affect the performance of MGHGA. In addition, there are differences in the way the hypergraphs of HGNN-KNN and HGNN-$\varepsilon$ are modeled making their accuracies in the classification task different, and the MGHGA performance differs under different combinations of HGNNs.

\section{Related Work}\label{S2}
\subsection{Hypergraph Learning}\label{HypergraphLearning}
The flexibility and capability of hypergraph learning to model complex higher-order data correlations have garnered increasing attention in recent years \cite{WOS:000853810100017,huang2021residual}. Hypergraph learning usually consists of two parts: constructing hypergraphs and designing hypergraph learning methods. (1) \textbf{Constructing hypergraphs}. There are four types of methods for constructing hypergraphs: distance-based, representation-based, attribute-based, and network-based. Specifically, Huang et al. \cite{huang2009video} proposed a nearest neighbor construction method whose main aim is to find adjacent vertices in the feature space and construct a hyperedge to connect them. Wang et al. \cite{wang2015visual} proposed a representation-based hyperedge construction mechanism that exploits the correlation between feature vectors to establish nodes connections. The literature \cite{huang2015learning} proposed a generation method applicable to attribute hypergraphs, which uses attribute information to construct hypergraphs. Fang et al. \cite{fang2014topic} used user friendship and mobility information to construct hypergraphs in the location social networks. (2) \textbf{Hypergraph learning methods}. Hypergraph learning can be divided into spectral analysis methods, neural network methods, and other methods according to their implementations. Feng et al.  \cite{feng2019hypergraph} first proposed the hypergraph neural network, which extends the spectral approach of graph convolutional neural networks to the hypergraph and designs hypergraph convolution operations. Yadati et al. \cite{yadati2019hypergcn} proposed the HyperGCN, which solves the problem of processing semi-supervised classification on the hypergraph. Huang et al. \cite{huang2021residual} proposed MultiHGNN, which learns multimodal hypergraph data and uses hypergraph modeling for each modality to accomplish downstream tasks. Jiang et al. \cite{jiang2019dynamic} proposed a dynamic hypergraph neural network, which consists of two modules: dynamic hypergraph construction and  convolution. Tran \cite{tran2020directed} proposed a directed hypergraph neural network based on the directed hypergraph Laplacian operator for the semi-supervised learning of the directed hypergraph.

\subsection{Graph Adversarial Attack}\label{2.2}
Graph attack algorithms can be classified into different types, mainly attack type, target, knowledge and capability. 
(1) Attacks can be classified into three categories based on their type: \textbf{the topology attack, the feature attack and the hybrid attack} \cite{FGA,GNAI,wu2019adversarial}. In the topology attack, the attacker focuses on modifying the graph topology, which is a common attack method, e.g., FGA \cite{FGA}, Mettack \cite{Mettack}, RL-S2V \cite{dai2018adversarial} and HyperAttack \cite{hu2023hyperattack}. The node feature modification is another common attack method, where the attacker focuses on modifying the features of the nodes, e.g., GANI \cite{GNAI}. In Nettack \cite{Nettack} and IG-Attack \cite{wu2019adversarial}, attackers use the graph topology and node feature attacks to degrade GNNs' accuracy.
(2) Based on the target of the attack, we can classify the attacks into the following two categories:\textbf{ targeted and untargeted attacks} \cite{dai2022targeted,2022arXiE220801819T}. Dai et al. \cite{dai2022targeted} proposed a targeted universal attack against GNNs, where the attacker's goal is to misclassify some of the test nodes into the attacker-specified labels. Fang et al. \cite{GNAI} injected fake nodes with malicious information into the graph which made the GNN perform very poorly on the test nodes. (3) According to the knowledge classification of the attacker can be divided into three categories: \textbf{the white box attack, the gray box attack and the black box attack} \cite{2022arXiv220812815L,Nettack}. In white box attack, the attacker knows all the knowledge about the GNNs model and datasets \cite{Mettack}. In a gray-box attack, the attacker only has some knowledge, e.g., knowing the parameters of GNNs but not the prediction results of nodes \cite{Nettack}. In black-box attacks, the attacker does not know the model architecture, parameters and training data, and can only obtain a small amount of model feedback \cite{dai2018adversarial}. Liu et al. \cite{2022arXiv220812815L} proposed a multi-level propagation surrogate white box attack where the attacker knows the model parameters and dataset information. The attack improved the success rate of the attack by querying the node information and using batch normalization to enhance the dissimilarity of node representations. Hussain et al.  \cite{hussain2022adversarial} proposed a gray-box attack where the attacker can access the labels of nodes and disrupt the fairness of node classification by injecting adversarial links. Ju et al. \cite{ju2022let} proposed a black-box attack method using a reinforcement learning framework, the attacker is not using a surrogate model to query model parameters or training labels. (4) Attacks can be classified into three categories based on the capabilities of the attacker: \textbf{the single node attack, the some node attack, and the all node attack} \cite{chen2022practical,2022arXiE220801819T}. Chen et al. \cite{chen2022practical} proposed a single node structure attack model proving that the single node attack can effectively reduce the accuracy of GNNs. Zang et al. \cite{zang2023guap} proposed a universal attack with modified edges in which the attacker reduces the effectiveness of GNNs by modifying a particular node or subgraph structure.

\section{Conclusion}\label{Sec:conclusion}
Our work shows that HGNNs are vulnerable to attacks in untargeted attack. In this paper, we present the first untargeted attack on HGNNs, named MGHGA. Considering the training differences between HGNN and GNNs, MGHGA uses surrogate models to modify node features before hypergraph modeling. Specifically, MGHGA uses the momentum gradient mechanism to select the features of the attack nodes. MGHGA uses different methods to update discrete and continuous features in the feature generation module. Extensive experimental results show that MGHGA can achieve advanced attack levels in node and visual object classification tasks.

In this paper, we only discuss the vulnerability of HGNNs. However, MGHGA has drawbacks. For example, MGHGA is set up as a white-box attack that accesses the HGNNs training parameters during the process of the attack. In some extreme cases, the attacker can only access some or none of the parameters, which leads to MGHGA failure. In our future work, we will consider two main aspects: (1) Consider the robustness of HGNNs in more scenarios, such as gray-box and black-box attacks. (2) According to the conclusion of this paper, we will consider how to improve the robustness of HGNNs under untargeted attacks.

There is a paucity of current research on the robustness of HGNNs. We hope that MGHGA is the first step in opening up exciting research avenues for studying HGNNs attacks and defenses.

\bibliography{sample.bib}
\bibliographystyle{named}

\end{document}